\documentclass[letterpaper, 10 pt, conference]{ieeeconf}  

\IEEEoverridecommandlockouts                              

\usepackage{graphics} 
\usepackage{epsfig} 
\usepackage{mathptmx} 
\usepackage{times} 
\usepackage{amsmath} 
\usepackage{amssymb}  

\usepackage[flushleft]{threeparttable}
\usepackage{makecell,booktabs}

\usepackage{color}

\usepackage{cite}
\usepackage{balance}
\usepackage{subfigure}
\usepackage{multirow}
\usepackage{array}
\usepackage{mathtools} 

\title{\LARGE \bf
Kinematic Parameter Optimization of a Miniaturized Surgical Instrument Based on Dexterous Workspace Determination
}

\author{Xin Zhi$^{\dag}$, Weibang Bai$^{\dag}$*,~\IEEEmembership{Member,~IEEE}, and Eric M. Yeatman,~\IEEEmembership{Fellow,~IEEE} 
\thanks{${\dag}$The authors contributed equally to this work, ${*}$the corresponding author. 
This work was supported by  EPSRC  project  EP/P012779/1.
}
\thanks{Xin Zhi is with Chongqing University-University of Cincinnati Joint Co-op Institute, Chongqing University, Chongqing, China.
}%
\thanks{Weibang Bai is with the  Hamlyn  Centre  and  the  Department  of  Computing, Imperial College London, Exhibition Road, SW7 2AZ, London, UK. 
{
wbbai@imperial.ac.uk.}}%
\thanks{E. M. Yeatman is with the Hamlyn Center and the Department of Electrical and Electronic Engineering, Imperial College London,  Exhibition Road, SW7 2AZ, London, UK.}%
}

\begin{document}

\maketitle
\thispagestyle{empty}
\pagestyle{empty}

\begin{abstract}
Miniaturized instruments are highly needed for robot assisted medical healthcare and treatment, especially for less invasive surgery as it empowers more flexible access to restricted anatomic intervention. But the robotic design is more challenging due to the contradictory needs of miniaturization and the capability of manipulating with a large dexterous workspace. Thus, kinematic parameter optimization is of great significance in this case. To this end, this paper proposes an approach based on dexterous workspace determination for designing a miniaturized tendon-driven surgical instrument under necessary restraints. The workspace determination is achieved by boundary determination and volume estimation with partition and least-squares polynomial fitting methods. 
The final robotic configuration with optimized kinematic parameters is proved to be eligible with a large enough dexterous workspace and targeted miniature size.

\end{abstract}

\section{INTRODUCTION}

Medical robots are widely used in recent years as they can benefit both patients and surgeons by improving the reachability and efficiency of surgical operations and healthcare treatments. Although medical robotics is popular and promising, there are still many challenges for the design, fabrication, integration and intelligent control as miniaturization, flexibility, dexterity, accuracy and safety are highly demanded in clinical application, especially for modern less and less invasive surgical interventions 
\cite{vitiello2012emerging, troccaz2019frontiers, le2016survey,bai2017modular}. 

However, miniaturized flexible robotic instruments with small size but large dexterous workspace are always difficult to achieve \cite{le2016survey}. As a result, besides the continuum robots\cite{burgner2015continuum}, small scale discrete structures based on serial \cite{hong2020design}, parallel \cite{baidevelopment,chen2018design}, and serial-parallel hybrid mechanisms\cite{wbai2015development,bai2017novel} are developed. Meanwhile, further miniaturization and optimization design are still on the way to build the next-generation medical robots. It should provide small enough size with enhanced flexible access to confined anatomical sites and increased dexterous manipulation for complex tasks. 
Thus, to determine the optimized structures of the miniaturized surgical robot that empower a maximum dexterous workspace, kinematic parameter optimization can be implemented.

In general, robotic kinematics can be firstly represented using the Denavit-Hartenberg (DH) convention \cite{siciliano2010robotics} with specific parameters and limitations included. To optimize the parameters and achieve large dexterous workspace, the basic mapping relationship between the workspace features and the kinematic parameters of the robot should be addressed or estimated. Normally, the robotic reachable workspace in Cartesian space can be represented using point clouds obtained from the forward kinematics with large amount of joint states by sampling methods like the Monte Carlo method \cite{rastegar1990manipulation}. For the determination of the dexterous workspace, many indices can be chosen with an acceptable threshold. For example, Kim et al. \cite{kim1991dexterity} introduced the order dependency and scale dependency problems of the measurement of robotic dexterity, and the relative manipulability measure 
was used to the robotic design.

To formulate the optimization problem relating the workspace and link parameters, the shape, boundary and volume of the workspace can be analyzed.
For the volume determination, geometric, analytical and numerical methods are commonly used\cite{merlet2006workspace}. It depends on the complexity of robot configurations and restrictions. For a planar manipulator with a 2D workspace, the exterior and interior boundaries can be found by analyzing joint limitations combing both geometric and analytical methods. Meanwhile, with the Monte Carlo method, the 2D workspace can be also estimated by indicating the end effector positions. To investigate the shape and volume of the workspace, the approximate boundary can be estimated with segmentation methods \cite{alciatore1994determining}, and Cao et al.\cite{cao2011engineering} also presented a method to generate the boundary curves by partitioning into a set of slices along certain axis.
The analytic mapping function between the workspace determination and the kinematic parameters are hard to derive. To achieve the parameter optimization, 
different numerical techniques like SQP and Genetic Algorithms for solving the optimizing problems have been mainly addressed \cite{panda2009appropriate, hosseini2011dexterous}. 

In this study, besides the popularly addressed solving techniques, the fundamental dexterous workspace determination study is also presented when developing a tendon-driven miniaturized surgical instrument for robotic assisted minimally invasive surgery. 
The overall size depends on the specific applications. 
In our case, it should be applicable for tissue manipulating surgical tasks like tumor resection, where primary tumor mean size 
is about 3cm \cite{koelliker2008axillary} which is similar to the size of a grape. For the distal articulated mechanism, the target of the design is to keep the structural size small and at least the total length should be less than a given value. Meanwhile, the manipulation with large enough dexterous workspace is also demanded. Therefore, the basic kinematic parameters of link lengths can be optimized.

\section{ROBOT MODELLING AND WORKSPACE DETERMINATION}

\subsection{Kinematic Modelling of the Miniaturized Surgical Robot}

According to the basic needs for the surgical instruments, the total link length should be acceptable to miniaturize the size while enlarging the dexterous workspace as much as possible. 
Normally, robotic surgical instruments need to have 6 Degrees of Freedom (DOF) for dexterous manipulating. Here, without considering the entire roll of the shaft, 
as it needs to be provided by the additional external actuation for this miniature tendon driven instrument,
the configuration for optimizing can be chosen with 5 DOF and set as PRRRR.

\begin{figure}[b] 
\vspace{-3mm}
    \centering
    \vspace{-2mm}
    \includegraphics[width = 0.4\textwidth]{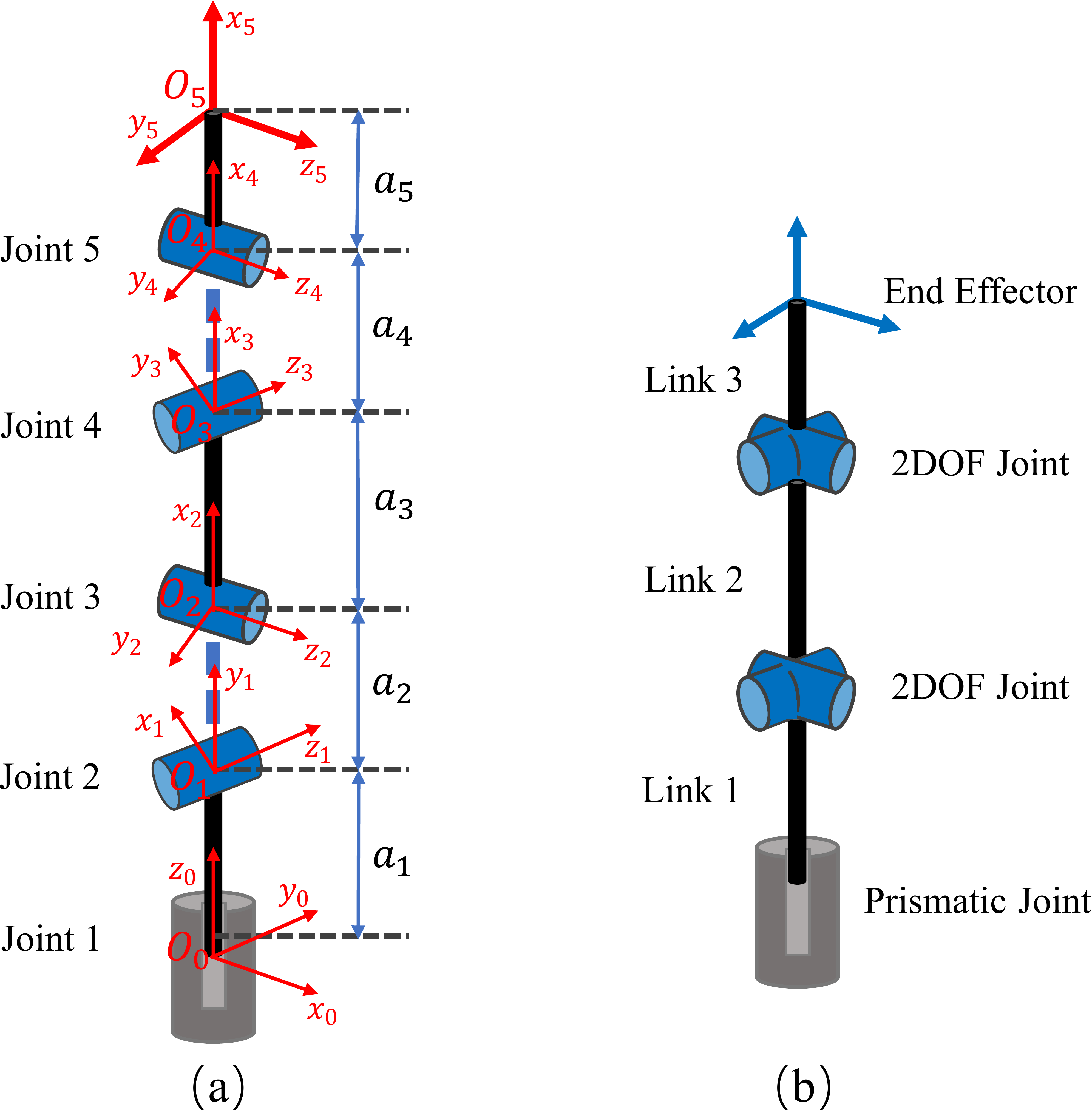}
    \vspace{-1mm}
    \caption{(a) General coordinate frames assignment \ (b) Actual robot configuration with two 2-DOF joints.}
    \label{Fig.Frames}
\end{figure}

The general kinematic chain of the derived 5 DOF serial surgical instrument can be described in Fig.\ref{Fig.Frames} (a) with coordinate frames of each joint independently assigned based on the DH convention. To make the robotic instrument more compact and miniaturized, we integrate the 5 DOF serial structure with one prismatic joint and two 2-DOF revolute joints in series. Then the actual kinematic configuration of initial state can be illustrated as Fig.\ref{Fig.Frames} (b). Meanwhile, the DH parameters can be specified with given feasible joint limits in Table.\ref{table:DHTable}, and accordingly we need to set the link length between the 2-DOF joints as zero. 
For miniaturization, the manufacturability and feasibility of integration are also of great importance for the design.
The minimal length of each link is set to be 3mm to make it fabricable, the last link will finally serve as end effector and would be longer than 10mm, and the total length of the robotic instrument is set to be less than 45mm. Thus, we have the basic considerations of the parameters like this:
\vspace{-1mm}
\begin{equation}
\left\{
\begin{array}{l}
a_2 = a_4 = 0 \\
a_{i} \geq 3,\quad i = 1,3\\
a_5 \geq 10 \\
L_{Total} = \sum_{i=1}^{5} a_i \leq 45
\end{array}
\right.
\label{equ:paralim}
\end{equation}

%
%
\begin{table}[t] 
\centering
  \caption{DH Parameters of the Miniaturized PRRRR Surgical Instrument}
  \vspace{-2mm}
  \begin{tabular}{c c c c c c c}
    \midrule\midrule 
    Link & $a_i/m$ & $\alpha_i/^\circ$ & $d_i/m$ & $\theta_i/^\circ$ & Offset & $\text{Joint Range}$  \\
    \midrule
    1 & $a_1$ & 90  & $d_1$ & $0$ & $90^\circ$ & $[0, 10mm]$ \\[0.5ex]
    2 & $a_2 = 0$ & -90 & $d_2 = 0$ & $\theta_2$    & $90^\circ$  & $[-45^\circ, 45^\circ]$   \\[0.5ex]
    3 & $a_3$ & 90  & $d_3 = 0$ & $\theta_3$ & $0$  & $[-45^\circ, 45^\circ]$  \\[0.5ex]
    4 & $a_4 = 0$ & -90 & $d_4 = 0$ & $\theta_4$   & $0$   & $[-45^\circ, 45^\circ]$  \\[0.5ex]
    5 & $a_5$ & 0 & $d_5 = 0$   & $\theta_5$   & $0$  & $[-45^\circ, 45^\circ]$  \\[0.5ex]
    \midrule\midrule
    \end{tabular}
    \label{table:DHTable}
    \vspace{-5mm}
\end{table}

Given the assigned frames and parameters, the tip position of the end effector $P{(X_e,\ Y_e,\ Z_e)}^T$ can be derived analytically by forward kinematics equation using the DH method:
\begin{equation}
\left\{
\begin{array}{l}
X_{e}=-a_{3} s_{3}-a_{5} c_{3} s_{5}-a_{5} c_{4} c_{5} s_{3} \\
Y_{e}=a_{3} c_{2} c_{3}-a_{5} c_{5}\left(s_{2} s_{4}-c_{2} c_{3} c_{4}\right)-a_{5} c_{2} s_{3} s_{5} \\
Z_{e}=a_{1}+d_{1}+a_{3} c_{3} s_{2}+a_{5} c_{5}\left(c_{2} s_{4}+c_{3} c_{4} s_{2}\right)-a_{5} s_{2} s_{3} s_{5}
\end{array}
\right.
\label{equ:posi}
\end{equation}
 where $c_i$ and $s_i$ represent $\cos(\theta_i)$ and $\sin(\theta_i)$ respectively.

\subsection{Workspace Analysis and Determination}

To further optimize the kinematic parameters of the derived robotic instrument for the targeted dexterous manipulation with given size of objects, the reachable workspace and dexterous workspace need to be analyzed. 

\subsubsection{Workspace Estimation}

\begin{figure}[ht]
    \centering
    \includegraphics[width = 0.29\textwidth]{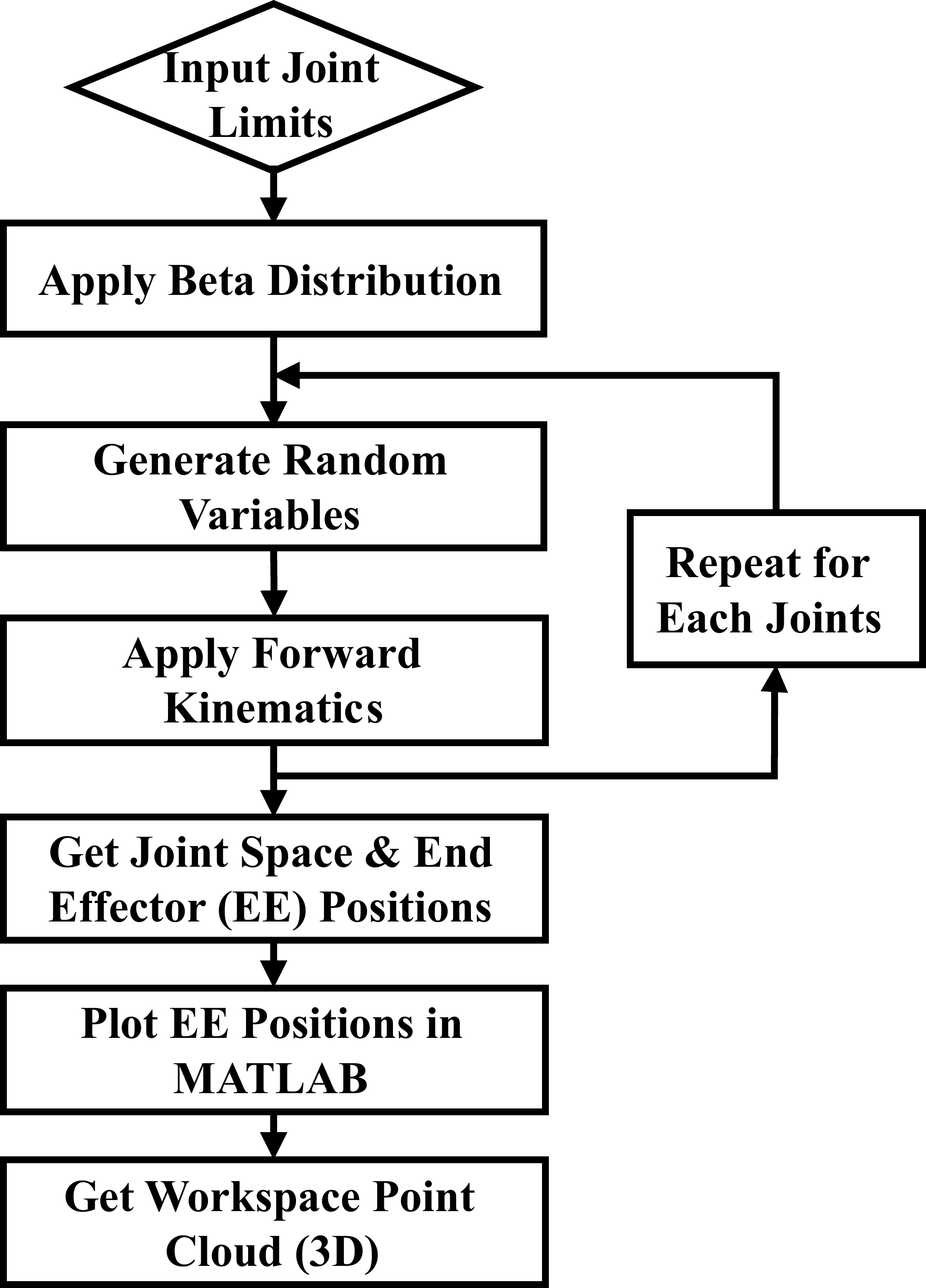}
    \caption{Schematic workflow of workspace estimation.}
    \label{Fig.AMCworkflow}
    \vspace{-5mm}
\end{figure}

The Monte Carlo method is a random sampling method normally used to numerically determine the workspace of robots. The basic idea is to randomly search various combinations of displacement and angle values within the joint limitations in joint space. Then the forward kinematics transformation of the robot is computed to get positions of the end effector. 
Thus, the robot workspace can be estimated and described by the derived point clouds in $\Re^3$ Cartesian  space. However, inaccuracy and nonuniform distribution problems 
often occur due to the nonlinearity of the transformation or mapping relationship between joint space and task space, which can be represented as $q \in \Re^N \mapsto X \in \Re^3$, here $N=5$. The advanced Monte Carlo method \cite{cao2011accurate, peidro2017improved} can be used to improve the accuracy of workspace boundary. The simplified schematic workflow of the workspace estimation based on the improved Monte Carlo method is presented in Fig.\ref{Fig.AMCworkflow}.

The joint state $q_i$ in joint space can be generated by Eq.\eqref{eq_jointValues} with the traditional Monte Carlo method:
\begin{equation} \label{eq_jointValues}
q_{i}=q_{i}^{\min }+\left(q_{i}^{\max }-q_{i}^{\min }\right) \operatorname{rand}(1,j)
\end{equation}
Where $i=1,\ldots,5$, represents the joint numbers, and $j$ is the number of randomly generated points. $\text{rand} \left(\right)$ is a normally used Matlab function uniformly generating $1 \times j$ random values within the interval $(0,1)$. 
Larger $N$ can increase the estimation accuracy but lead to longer time for calculating. We set $N = 10,000$ to ensure an acceptable accuracy.

To compensate for the nonuniform density distribution problem of traditional Monte Carlo method, a modification technique called Beta random distribution model \cite{cao2011accurate} is used.
Beta random distribution model involves a variety of shapes depending on two parameters, $\alpha$ and $\beta$. By selecting the values of $\alpha$ and $\beta$, the distribution of end effector locations in Cartesian space will achieve higher uniformity. In this study, the values of $\alpha$ and $\beta$ are determined by Eq.\eqref{eq:Betadist}: 
\begin{equation}\label{eq:Betadist}
\alpha_{i}=\beta_{i}=\frac{\Delta\vartheta_{i}}{5 \pi}+0.30
\end{equation}
Where $\Delta \vartheta_i$ is the limit range of each revolute joint, i.e. $\Delta \vartheta_i = q_i^{max}-q_i^{min}$. For the prismatic joint, the same shape of Beta density function is used. Then, the joint state $q_i$ in \eqref{eq_jointValues} is updated as:
\begin{equation} \label{eq_jointValuesmod}
q_{i}=q_{i}^{\min }+\left(q_{i}^{\max }-q_{i}^{\min }\right) \cdot \operatorname{betarnd}\left(\alpha_{i}, \beta_{i},1,j \right)
\end{equation}
Where $\operatorname{betarnd}\left(\alpha_{i}, \beta_{i},1, j \right)$ is the Matlab function for generating $1 \times j$ random numbers from the beta distribution with $\alpha_i$ and $\beta_i$.

With forward kinematics computing, we can obtain the mapping function from joint states ${q}=[q_1… q_5]^T$ 
to the locations of the end effector  ${P}_{e}=[X_e, \ Y_e, \ Z_e]^T$, as indicated in the analytical form in \eqref{equ:posi}. Meanwhile, the Cartesian space positions of the end effector can be visualized with 3D point clouds in Matlab. 

\subsubsection{Dexterous Workspace Determination}
Dexterous workspace of a robot defines the spacial positions that can be reached in any orientation by the end effector,
which indicates the dexterous manipulating performance.
The size and volume of the dexterous workspace are heavily affected by the constraints of the joint space and the robotic configurations. 

To optimize the kinematic design of the PRRRR surgical instrument and achieve best dexterity for given surgical tasks of manipulating a piece of tissue with defined size, the relative manipulability measure $M_r$ which is dependent of the order of both scale $L_{Total}$ and task space $m~(\text{which specifies}~J\in \Re ^{m\times n})$ is adopted \cite{kim1991dexterity}. The manipulability $M$ is also deduced and modified based on the traditional definition in \cite{yoshikawa1985manipulability}.

\begin{equation}
M_{r}=\frac{M}{f_{M}}
\label{equ:mr}
\end{equation}
\begin{equation}
M=\sqrt[\uproot{6}m]{\operatorname{det}\left(J J^{T}\right)}
\label{equ:m}
\end{equation}

In Eq.\eqref{equ:mr}, $f_M$ is a function of the dimension of the robot total length, and can be simply defined as the total link length in this case, which would be $f_M = L_{Total} = a_1 + a_3 + a_5$. 
%

\begin{figure}[htp]
\centering 
\vspace{-1mm}
\subfigure[Overall view] {
    \label{fig.WS_a}     
    \includegraphics[width=0.49\columnwidth]{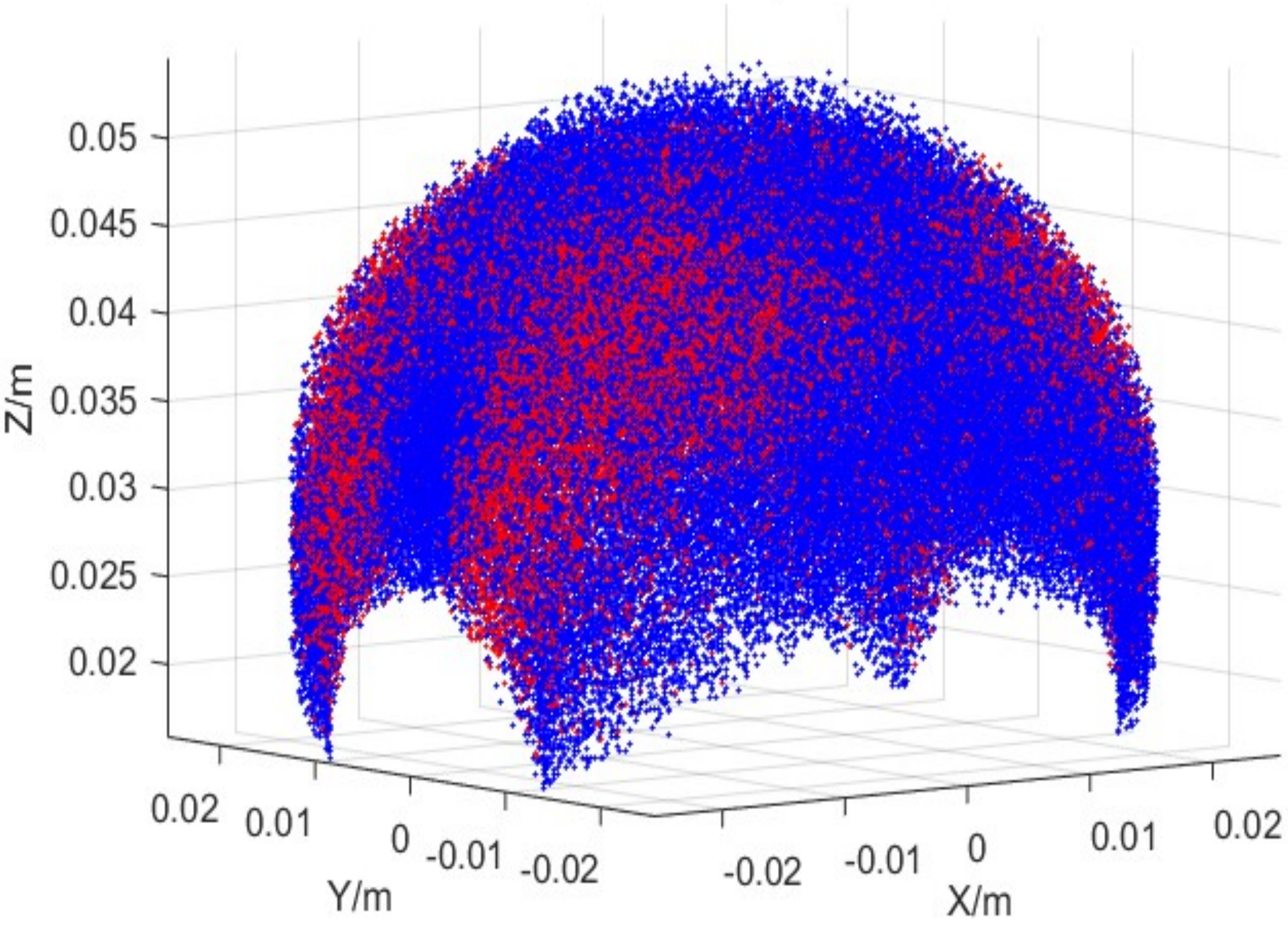}  
}     
\subfigure[Side view] { 
    \label{figWS_b}     
    \includegraphics[width=0.44\columnwidth]{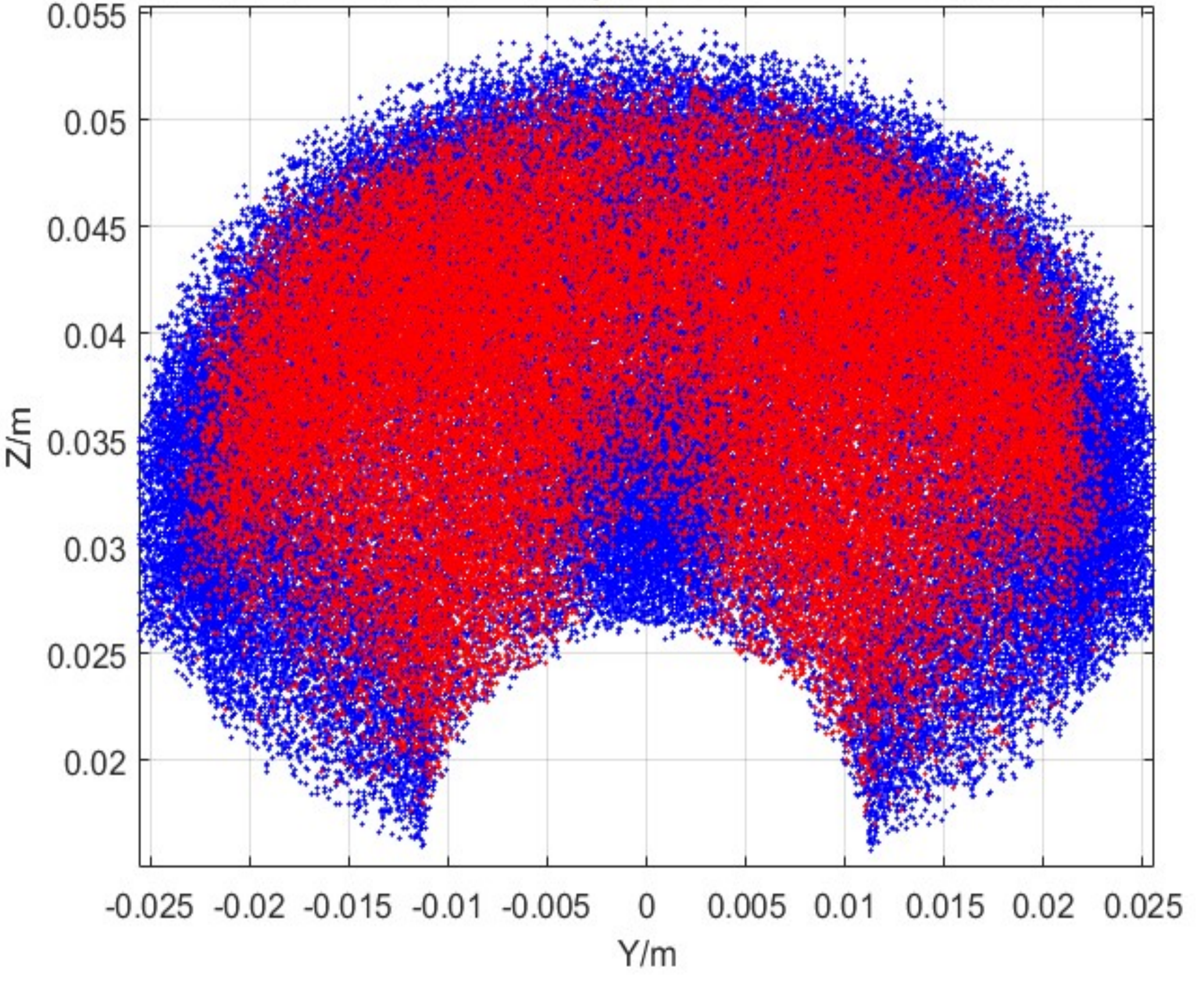}     
}

\caption{Reachable (blue) and dexterous (red) workspace representation. }  
\label{fig.WSab} 
\vspace{-2mm}
\end{figure}

In comparison to the traditional manipulability measure of $M=\sqrt{\operatorname{det}\left(J J^{T}\right)}$ defined in \cite{yoshikawa1985manipulability}, which results in an index of dexterity of the order of magnitude of ${10}^{-11}$ in this case,
the relative manipulability measure $M_r$ will get the results at the order of magnitude of ${10}^{-1}$. 
To determine the acceptable points in dexterous workspace, an acceptable value $m_{ref}$ for the calculated $M_r$ can be specified. Here we can define $m_{ref} = 65\%$, which means the index larger than 65\% of the total range is recognized as the acceptable dexterous workspace. The obtained dexterous workspace is represented by red points and reachable workspace by blue points, as shown in Fig.\ref{fig.WSab}. 

\subsubsection{Workspace Boundary Determination}

\begin{figure}[htp]
\centering 
\vspace{-1mm}
    \includegraphics[width = 0.49\textwidth]{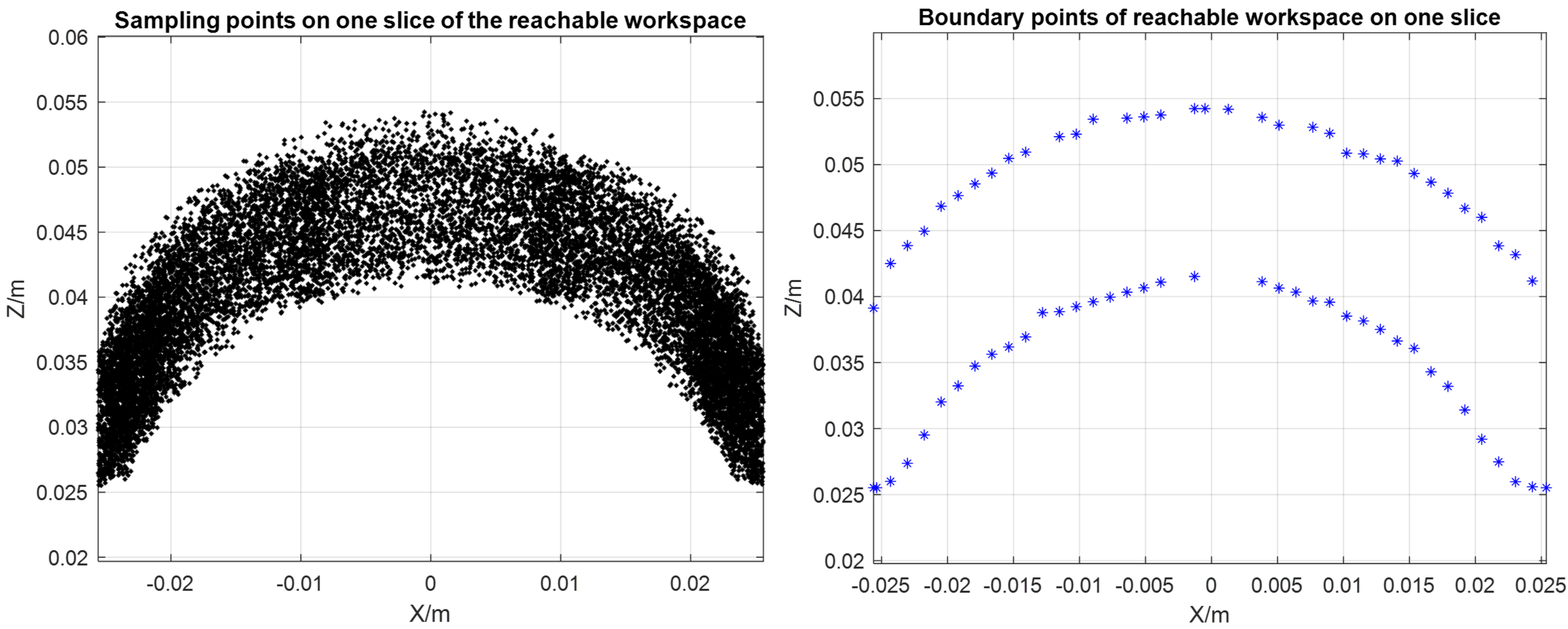}
\vspace{-4mm}
\caption{Sampled points and the upper and lower boundary of one partitioned slice of the obtained reachable workspace.}  
\label{fig.WSslicebound}   
\vspace{-2mm}
\end{figure}

To further depict the shape and profile of the obtained workspace with resulting point clouds, the boundary can be first analyzed. By partitioning the obtained approximate workspace into $S_n$ slices along the $Y$ direction, the specific shape of each can be indicated. Then along the $X$ direction of the resulting slices of the 2D point cloud, it can be further divided into $S_m$ different columns to figure out the upper and lower boundary points, i.e. the maximum $Z_{\max}(n_i,m_j)$ and minimum $Z_{\min}(n_i,m_j)$ at $n_i$ slice $m_j$ column, where $n_i = 1, \cdots, S_n$, $m_j = 1, \cdots, S_m$. Take one slice as an example, the results can be seen in Fig.\ref{fig.WSslicebound}. 
In this way, the boundary of each slice and thus the overall workspace can be figured out. 

If the partition is evenly performed, the width $\delta_y$ between each slice can be calculated by: 
\begin{equation}
	\delta_{y}=\frac{Y_{\max }-Y_{\min }}{S_n}
\label{eq.Zslice}
\end{equation}
where $Y_{max}$,\ $Y_{min}$ are the maximum and minimum $y$ values of the estimated workspace, $S_n$ is the number of slices and set to be 40 in this study.

Similarly, for the columns along $X$ direction, we have:
\begin{equation}
    \delta_{x}=\frac{X_{\max }-X_{\min }}{S_m}
\end{equation} 
where $X_{\max}$,\ $X_{\min}$ are the maximum and minimum $x$ values of the sampled points within each slice, $S_m$ is the number of columns for one slice and thus  $S_m$ pairs of points will be generated for each.

\begin{figure}[htp]
\centering    
\subfigure[Boundary points and resulted fitting curves with $S_m = 400$] {
    \label{fig.WS_slicbound400}     
    \includegraphics[width=0.95\columnwidth]{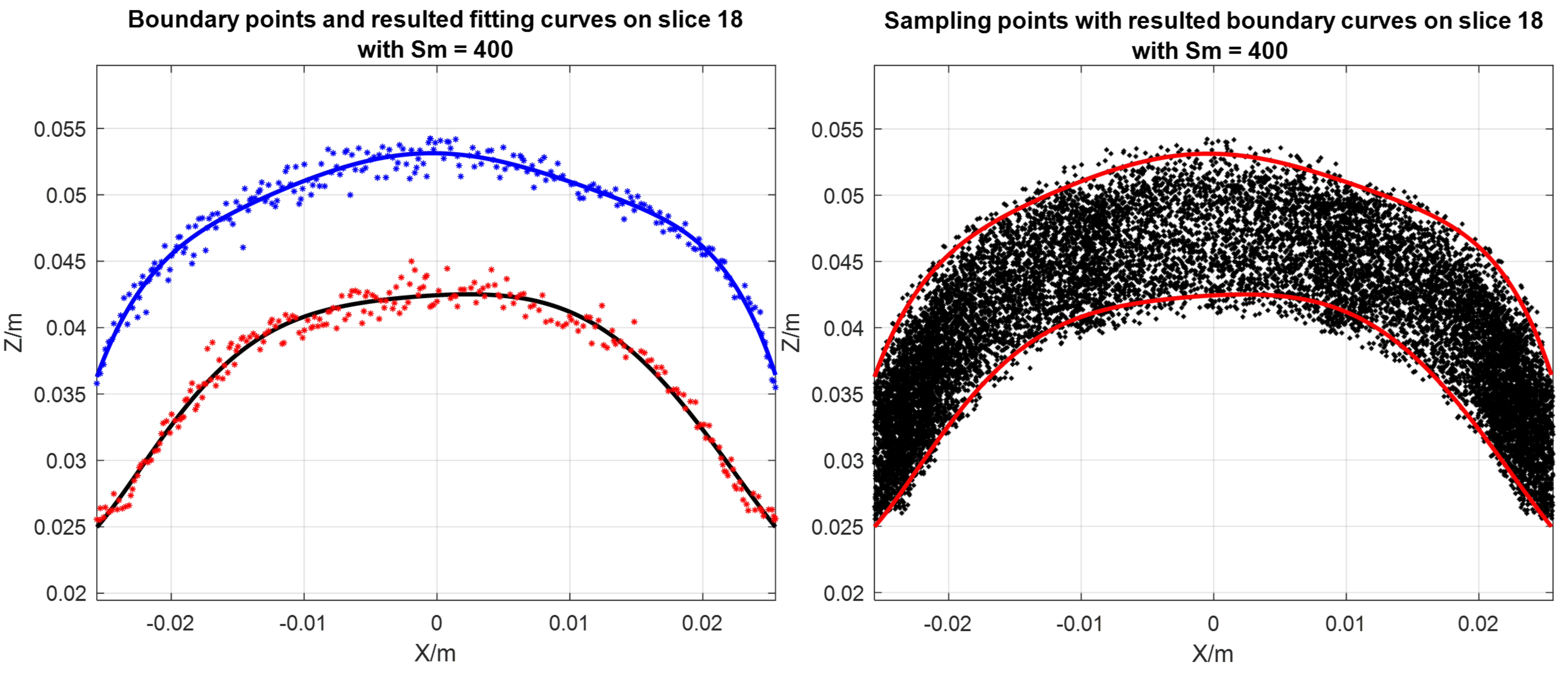}  
}     
\subfigure[Boundary points and resulted fitting curves with $S_m = 40$] { 
    \label{fig.WS_slicbound40}     
    \includegraphics[width=0.95\columnwidth]{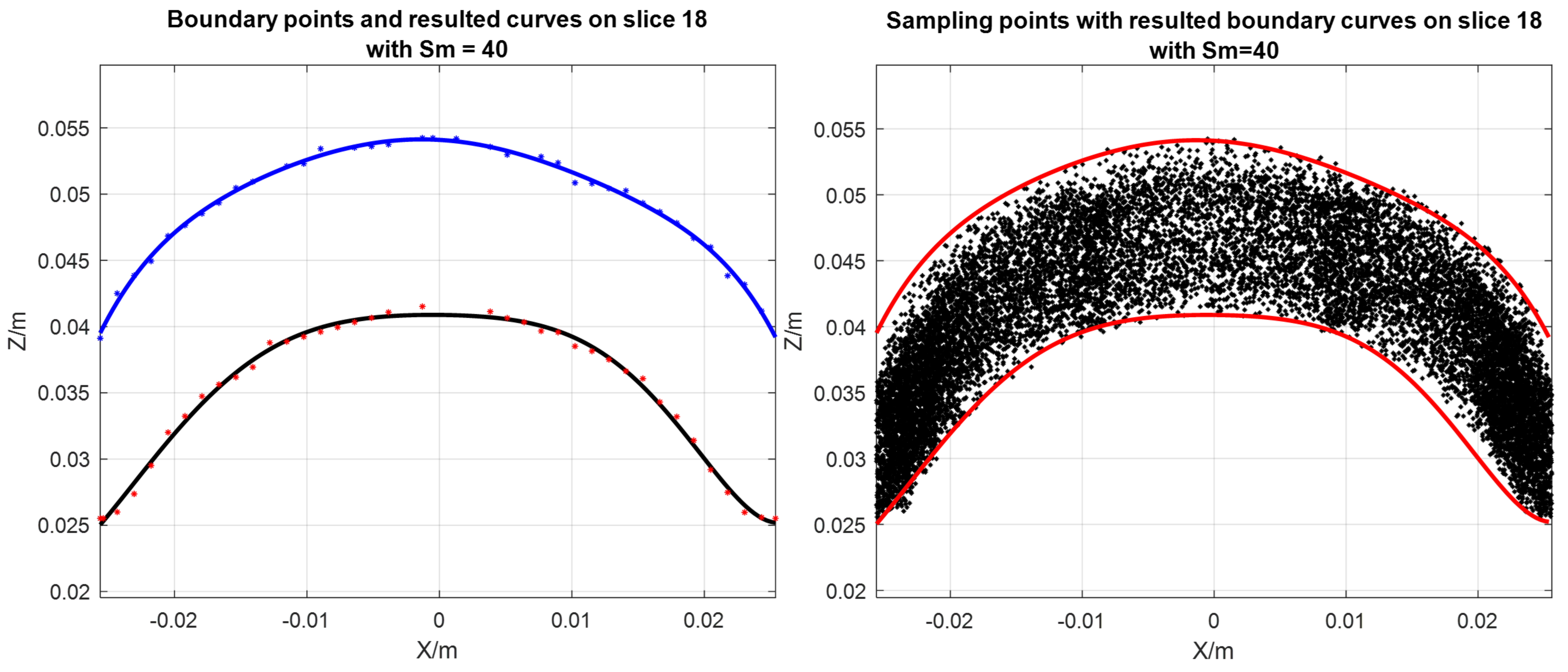}     
}
\vspace{-1mm}
\caption{Comparison of the boundary determination with different number of columns $S_m$ on selected 18th slice given 500,000 sampling points in total. }  
\label{fig.WS_slicboundComp}   
\vspace{-3mm}
\end{figure}

With the boundary points of each slice, the cross-sectional shape can be specified by curve fitting method with polynomial functions. But according to the tests,  it is found that the number of points which is decided by the columns $S_m$ for each slice will greatly affect the results of the fitting curve. In contrast, when setting different $S_m$ as shown in Fig.\ref{fig.WS_slicboundComp}, it is obvious that the resulted boundary accuracy in Fig.\ref{fig.WS_slicbound40} with  
$S_m = 40$ is much better than the results of Fig.\ref{fig.WS_slicbound400} with $S_m = 400$. More pairs of boundary points may lead to over-fitting problems for this problem.

Thus, to get better boundary representation and reduce the fake points or noise for defining the real boundary as accurately as possible, the $\delta_x$ should be chosen properly. After a few trials, for the obtained point clouds formed with 500,000 sampling points, the number of columns for each slice along $X$ axis $S_m$ is set to be 40 in this study.

With the pairs of boundary points for each column on each slice $\left( Z_{\max}(n_i,m_j), \ Z_{\min}(n_i,m_j) \right)$, we can specify the upper boundary curve $u\left(x\right)$ and the lower boundary curve $l\left(x\right)$ using polynomial curve fitting methods. For the $n_i$ layer:
\begin{align}
    u(x)_{n_i}=p_{su}^{u} x^{su}+\cdots+p_{1}^{u} x^{1}+p_{0}^{u} 
    \label{eq:ux}\\
    l(x)_{n_i}=p_{sl}^{l} x^{sl}+\cdots+p_{1}^{l} x^{1}+p_{0}^{l}
    \label{eq:lx}
\end{align}
Where $p_0^u$, \ldots, $p_{su}^u$ and $p_0^l$, \ldots, $p_{sl}^l$ are coefficients of the upper and lower boundary polynomials, which can be solved by the Least Squares Method 
, and $su$ and $sl$ are consistent with the fitting order for the different boundary curves.

\begin{figure}[htp]
    \centering
    \includegraphics[width=0.93\columnwidth]{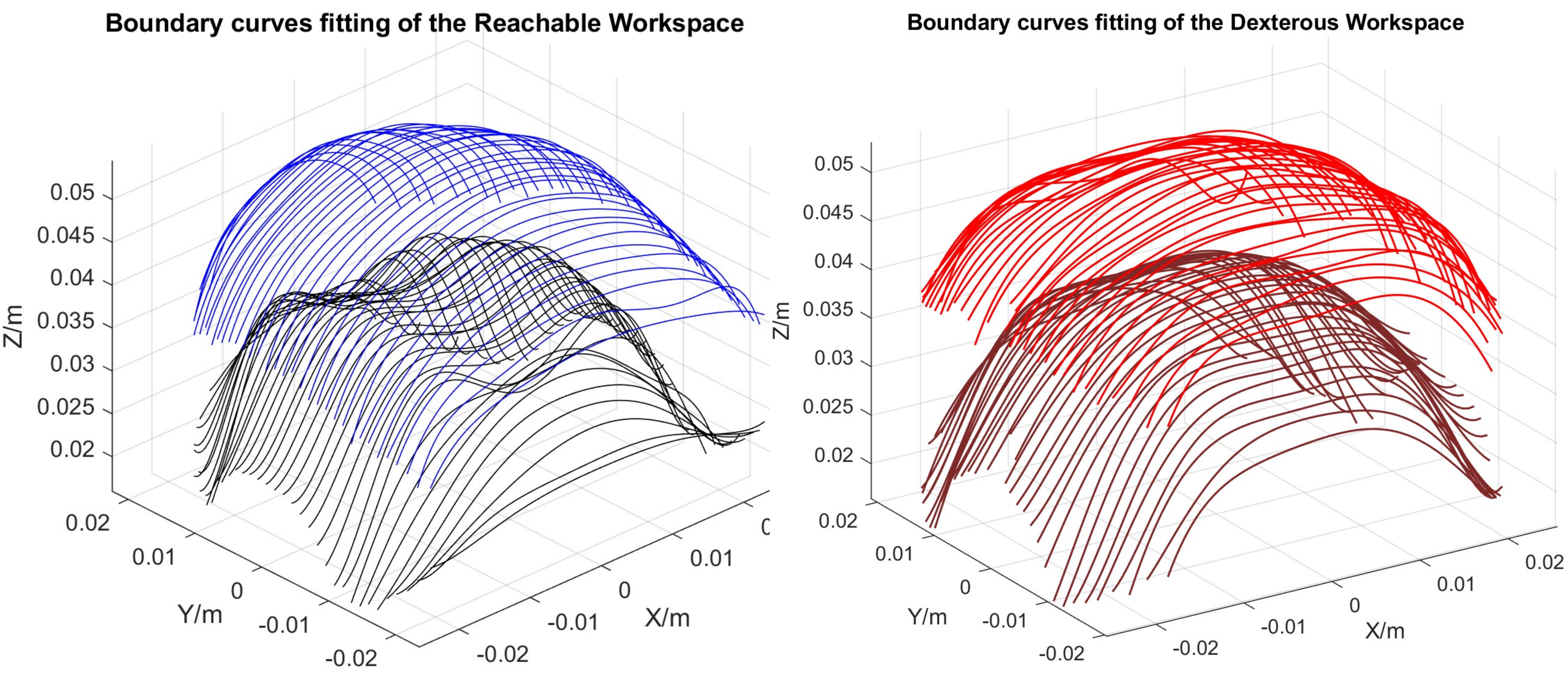} 
    \vspace{-1mm}
    \caption{Boundary curve fitting for reachable and dexterous workspace.}
    \label{fig.boundcurve}
\vspace{-2mm}
\end{figure}
To simplify the issue, we set the order of the polynomial functions for each slice as the same $su = sl = s$, and set $s = 7$ for the reachable workspace curve fitting and set $s = 6$ for the dexterous workspace. The boundaries of reachable workspace and dexterous workspace are shown in Fig.\ref{fig.boundcurve}, where the blue and black curves are upper and lower boundaries for the reachable workspace and the red and brown curves are upper and lower boundaries for the dexterous workspace.

\subsubsection{Workspace Volume Determination}

\begin{figure}[thpb]
    \centering
    \includegraphics[width = 0.29\textwidth]{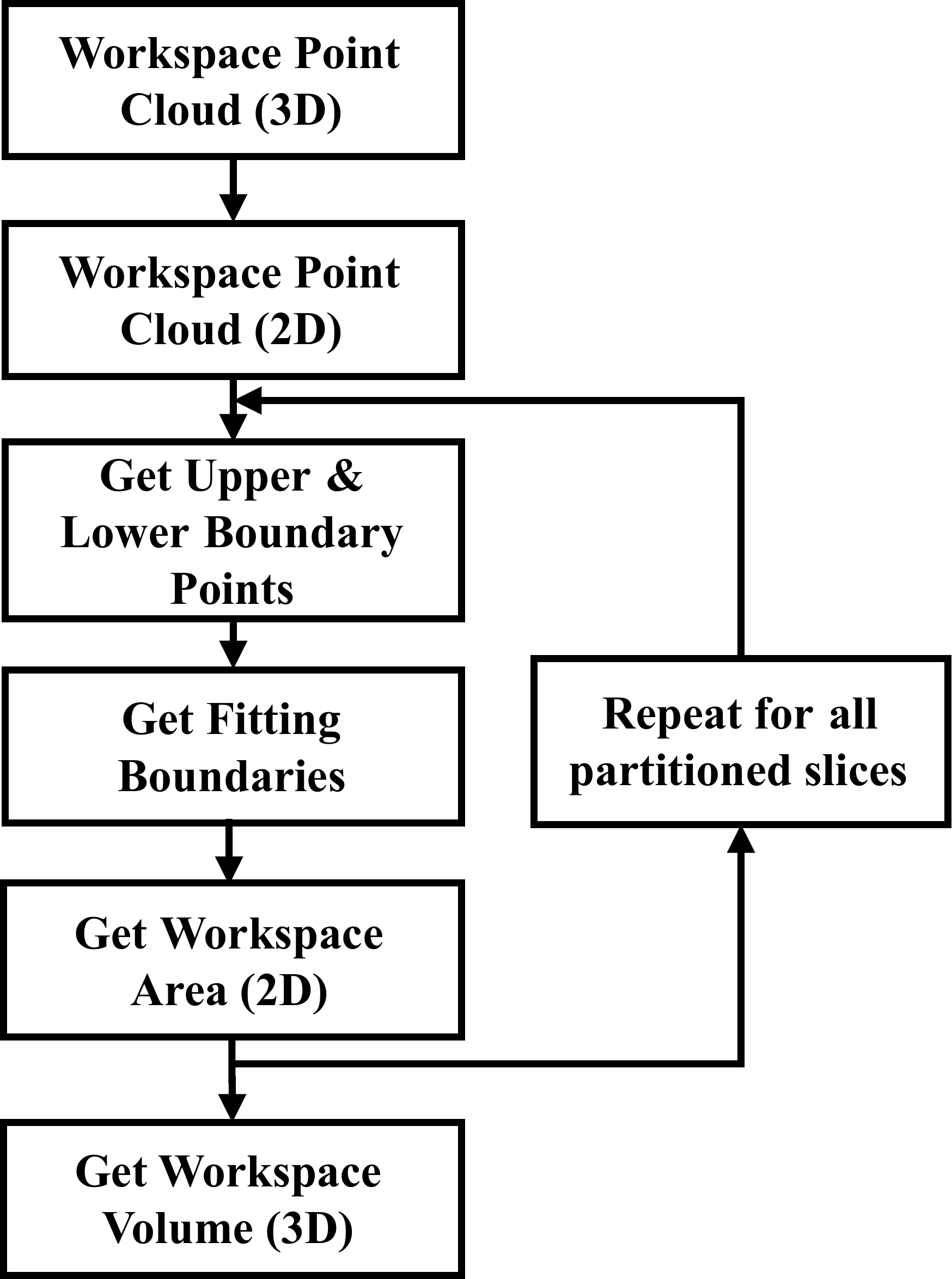}
    \vspace{-1mm}
    \caption{Schematic workflow of workspace volume determination.}
    \label{fig.WSvolumeflow}
\vspace{-3mm}
\end{figure}

In this work, a combination of numerical and analytical method is adopted to calculate the volume of the deduced PRRRR robot workspace. The schematic working principles the combination is summarized in Fig.\ref{fig.WSvolumeflow}.

Volume dimensional degradation
numerical integration method [9] is used to estimate the workspace volume. The generated workspace is first divided into $S_n$ slices along the $y$ axis with the width of $\delta_y$. Each slice is considered to be a 2D workspace, and the calculation of total volume is simplified by converting the problem from 3D domain to 2D domain. Consequently, the calculation time is reduced.

The area of each 2D workspace can be calculated by both linear approximation methods such as rectangular and trapezoidal, and non-linear approximation methods like parabolic and spline curves.
The commonly used rectangle-based numerical method is simple but will lead to large errors when the workspace has a void inside the shape and the absolute slope of the workspace boundary increases \cite{cao2011accurate}.
Unfortunately, according to the depicted shape of the PRRRR robot workspace, void space exists in the middle and its size is non-negligible comparing to the overall workspace. Thus, a more accurate approximation method is needed to calculate the 2D workspace area. 

The boundary curves in each slice were obtained using the least-squares $n$th order polynomial fitting method in previous section. The 2D workspace can be described by an envelope of an upper and a lower boundary curve. Thus, the area of the shape can be calculated by:
\begin{equation}
    \text {A}=\int_{X_{\min }}^{X_{\max}}[u(x)-l(x)] d x
    \label{eq:An}
\end{equation}
By applying Eq.\eqref{eq:ux} and Eq.\eqref{eq:lx} into Eq.\eqref{eq:An}, we can get the area of the 2D workspace:
\begin{equation}
\begin{small}
\begin{aligned}
     A = &\int_{X_{\min }}^{X_{\max }} \left[ \left(p_{s}^{u} x^{s}+\cdots+p_{1}^{u} x^{1}+p_{0}^{u}\right) - \left(p_{s}^{l} x^{s}+\cdots+p_{1}^{l} x^{1}+p_{0}^{l} \right) \right] dx \\
     = & \left(\frac{\Delta p_{s}}{s+1} X_{\max}^ {s+1}+\cdots+\frac{\Delta p_{1}}{2} X_{\max}^{2}+\Delta p_{0} X_{\max}^{1}\right) - \\
     & \left(\frac{\Delta p_{s}}{s+1} X_{\min} ^{s+1}+\cdots+\frac{\Delta p_{1}}{2} X_{\min} ^{2}+\Delta p_{0} X_{\min} ^{1}\right)
\end{aligned}
\end{small}
\end{equation}
where $\Delta p_k = p_k^u - p_k^l$, $k = 0, \cdots, s$.

As the obtained 2D workspace in each slice is symmetric, we can calculate it by:
\begin{equation}
    \begin{aligned}
    A=& 2\left(\frac{\Delta p_{s}}{s+1} X_{\max} ^{s+1}+\cdots+\frac{\Delta p_{1}}{2} X_{\max} ^{2}+\Delta p_{0} X_{\max} ^{1}\right)\\
    =& 2\sum_{j=1}^{s+1} \frac{\Delta p_{j-1}}{j} X_{\max}^{j}
    \end{aligned}
\end{equation}
Noticing that, the polynomial fitting coefficients $\Delta p_s \cdots \Delta p_0$ are varying with different slices, $X_{\max}$ and $X_{\min}$ are also different for each slice. 

Thus, the volume of a workspace $\Omega$ can be approximated by the summation of all the 2D workspace area $A(i) (i=1, \ldots, S_n)$ times the uniformly divided width $\delta_y$: 
\begin{equation}
\begin{aligned}
    V(\Omega) =&\delta_{y} \sum_{i=1}^{S_n} A(i) \\
    =&2\left(\frac{y_{\max,i }-y_{\min,i }}{S_n}\right) \sum_{i=1}^{S_n} \sum_{j=1}^{s+1} \frac{\Delta p_{j-1}^{i}}{j} X_{\max , i}^{j}
\end{aligned}
\label{eq:ws_volume}
\end{equation}

Nevertheless, the accuracy of this method is limited by the number of partitioned slices along $Y$ axis and the estimated 2D workspace areas $A(i)$, and larger $\delta y$ will lead to larger errors.

\section{PARAMETER OPTIMIZATION}

Robot kinematic parameter optimization is important for achieving better performance with surgical intervention tasks. With the basic limitation for each joint motion, the feasible length of each link can be optimized to achieve better dexterity for the end effector.

In this study, the specified PRRRR robotic instrument needs to be miniaturized in scale but also needs to achieve larger dexterous workspace.

Given all the limits of the parameters, the target can be simplified to find the best combination of each link length so that the PRRRR surgical instrument 
can achieve enough dexterous workspace for manipulating a tissue object about the size of a grape. The basic configuration of the instrument is determined and the parameter we can optimize in this case is the link length vector $\xi = [a_1, a_3, a_5]$.

For the specific location of the end effector $X = [x, y, z]$ inside the workspace $\Omega$, i.e. $X \in \Omega$, we can easily get the analytic mapping function with robotic forward kinematics expressions and reorganize it like: $X = f(\xi)$. However, the volume $V$ of the obtained workspace represented with the sampled 3D point clouds is hard to be addressed analytically with these kinematic parameters, especially when the workspace is geometrically non-symmetric. As a result, the volume $V(\Omega)$ is estimated based on the boundary determination and area accumulation strategy, as expressed in \eqref{eq:ws_volume}, but it is not providing us the analytic mapping function  of $V(\Omega) = g(\xi)$. Therefore, it is not easy to directly choose a pertinent solver for this optimization problem.

\begin{figure}[h]
    \centering
    \vspace{-2mm}
    \includegraphics[width = 0.33\textwidth, height = 0.51\textheight]{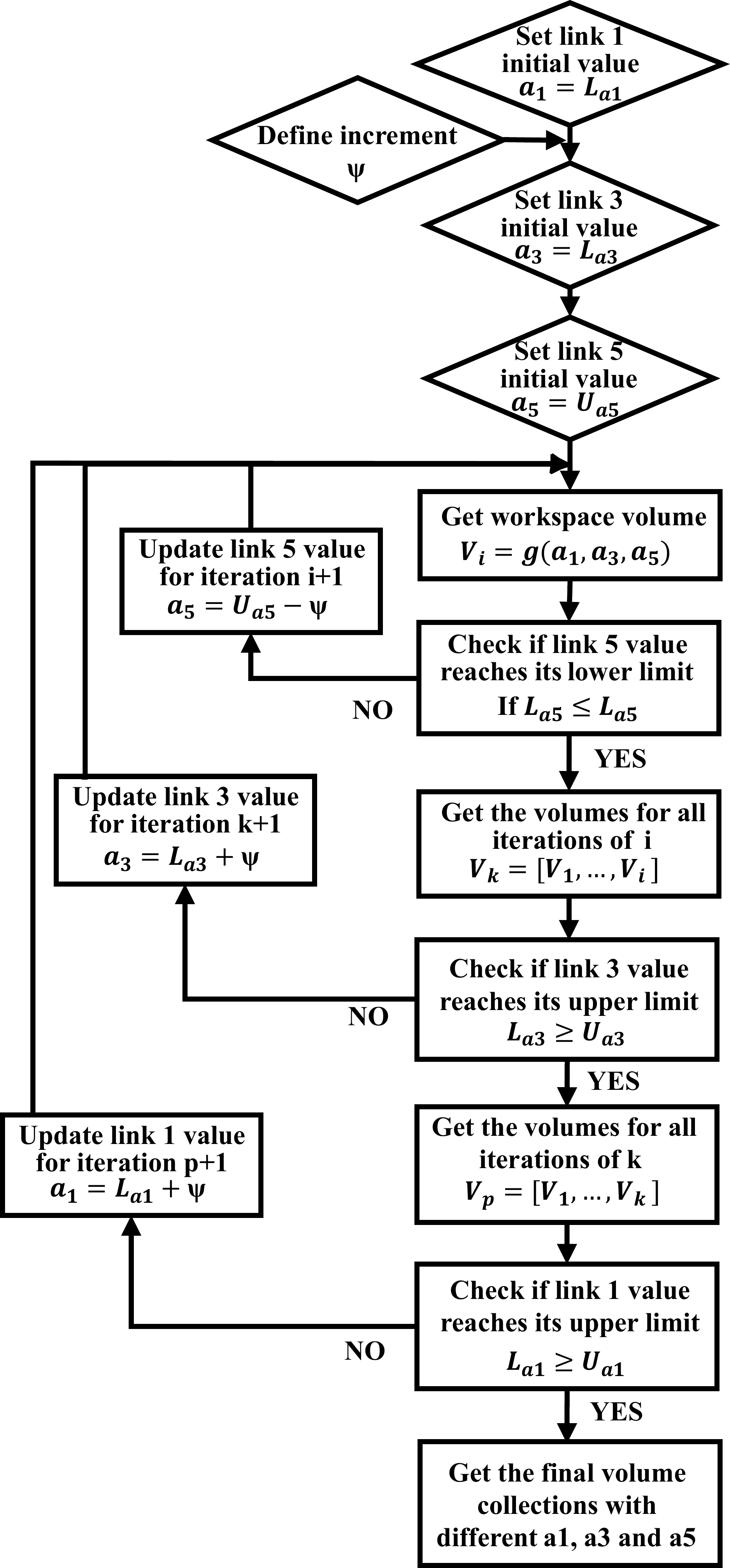}
    \vspace{-1mm}
    \caption{Illustration of the full exploration algorithm for the link length optimization based on the workspace determination.}
    \label{fig.Optworkflow}
\end{figure}
On the other hand, we only have a few link length variables to optimize in this case, so we can analyse the effect on the workspace volume by each link parameter numerically instead of addressing all the multiple parameters at the same time. Like in\cite{tsai1984effect}, the effect of each link length can be analyzed.
Thus, we can design the road map for solving this optimization problem as illustrated in Fig.\ref{fig.Optworkflow}. Where the feasible range of $a_1$ and $a_3$ are defined as $Range_{a_1} = [L_{a_1}, U_{a_1}]$, $Range_{a_3} = [L_{a_3}, U_{a_3}]$, and $Range_{a_5} = [L_{a_5}, U_{a_5}]$. 

The joint motion limits are structurally restricted and set to be within $[-45^\circ,\ 45^\circ]$ accordingly. The link length restrictions should guarantee the fabricability and basic functionality needs as indicated in \eqref{equ:paralim}. Thus, we can specify the feasible ranges as: $Range_{a_1} = [3, 8]mm$, $Range_{a_3} = [3, 12]mm$, and $Range_{a_5} = [15, 35]mm$.

Defining $\psi = [\psi_1, \psi_3, \psi_5]^T$ as the length increment for each link, and according to the computing complexity and the resolution of the fabricability of the miniature structures, we can set it to be $0.5mm$ or $1 mm$. 
To make it easier, we choose the increment as $\psi_1 = \psi_3 = \psi_3 =1mm$.
Given initial value of  $\xi = [a_1, a_3, a_5]$, during the iteration $p, \ k \ \text{and} \ i$, we can get all the possible combinations of the link parameters and their resulted dexterous workspace determination and estimation.

During each iteration, $\xi$ can be updated by:
\begin{equation}
\left\{
\begin{aligned}
    a_{1}=L_{a_{1}} + \psi_1 \\
    a_{3}=L_{a_{3}} + \psi_3 \\
    a_{5}=U_{a_{5}} - \psi_5 
\end{aligned}
\right.
\end{equation}
Then the maximum of the dexterous workspace volume ${V}_{max}$ can be updated iteratively until the final result is obtained.

\section{RESULTS}
\begin{figure}[thpb]
    \centering
    \includegraphics[width = 0.49\textwidth]{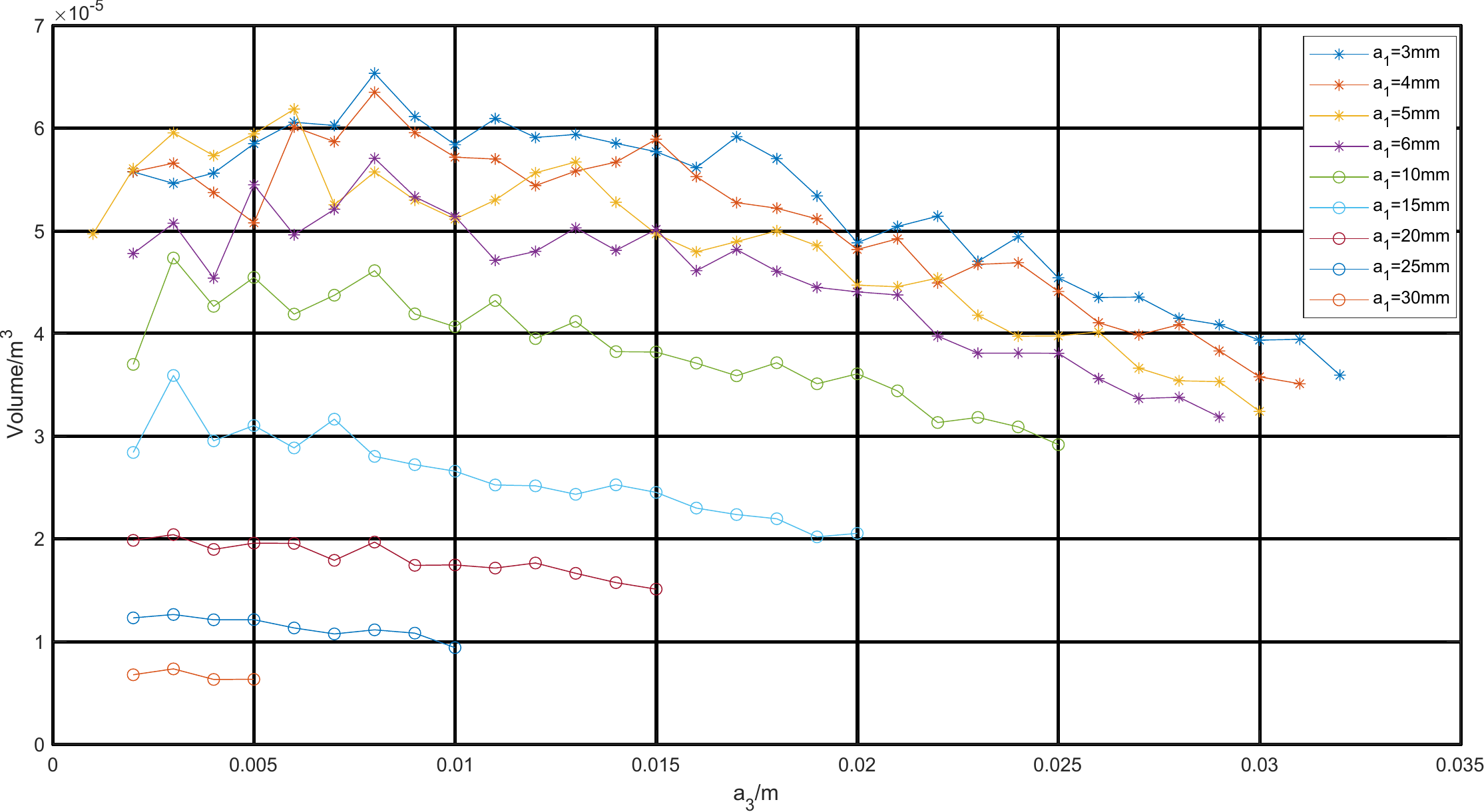}
    \caption{Results of the simplified exploration with selected  combinations of the link parameters.}
    \label{fig.allparavolume}
    \vspace{-5mm}
\end{figure}

Firstly, simplification process can be implemented for the optimization problem to get primary results less time-consumingly. According to the numerical calculating progress, we can find the link length $a_1$ has a monotonically decreasing effect on the resulted ${V_{k}}^{max}$, and the longer length of the last link $a_5$ will increase the total workspace with given joint limits for each. Thus, to preliminarily obtain the basic optimization results, the $L_{Total}$ can be set to be the maximum, i.e. $L_{Total}=45mm$, and $a_5$ can be determined by given the exploration of $a_1$ and $a_3$. Then the computing workflow in Fig.\ref{fig.Optworkflow} will be simplified, and 

with a few explorations of the link parameters within the limits, we can quickly get the distribution of the main results, as shown in Fig.\ref{fig.allparavolume}. According to the results of $V_{max}$, we can find the best combination of link parameters are: $a_1=3mm,\ a_3=8mm$, and $a_5=34mm$, with a maximum dexterous workspace volume of $7.69\times{10}^{-5}\ m^3$, and the volume of the reachable workspace in this case is $8.11\times{10}^{-5}\ m^3$. 

To further analyze the resulting robot workspace, the equivalent radius is computed to present an intuitional evaluation of the results. The equivalent spherical radius can be estimated by $R_e =\sqrt[3]{{3V}/{4\pi}}$. Consequently, the result of equivalent radius for the dexterous workspace is $R_{ed}=26.4mm$. This result indicates that the dexterous workspace of the optimized PRRRR robot instrument is similar to a sphere with a radius of $26.4 mm$, which is much larger than the defined target of $3cm$ in diameter as introduced in the beginning, and obviously it can be used in robotic surgery for manipulating the tissue object with defined scale as large as a grape. 

\begin{figure}[thpb]
    \centering
    \vspace{-5mm}
    \includegraphics[width = 0.45\textwidth]{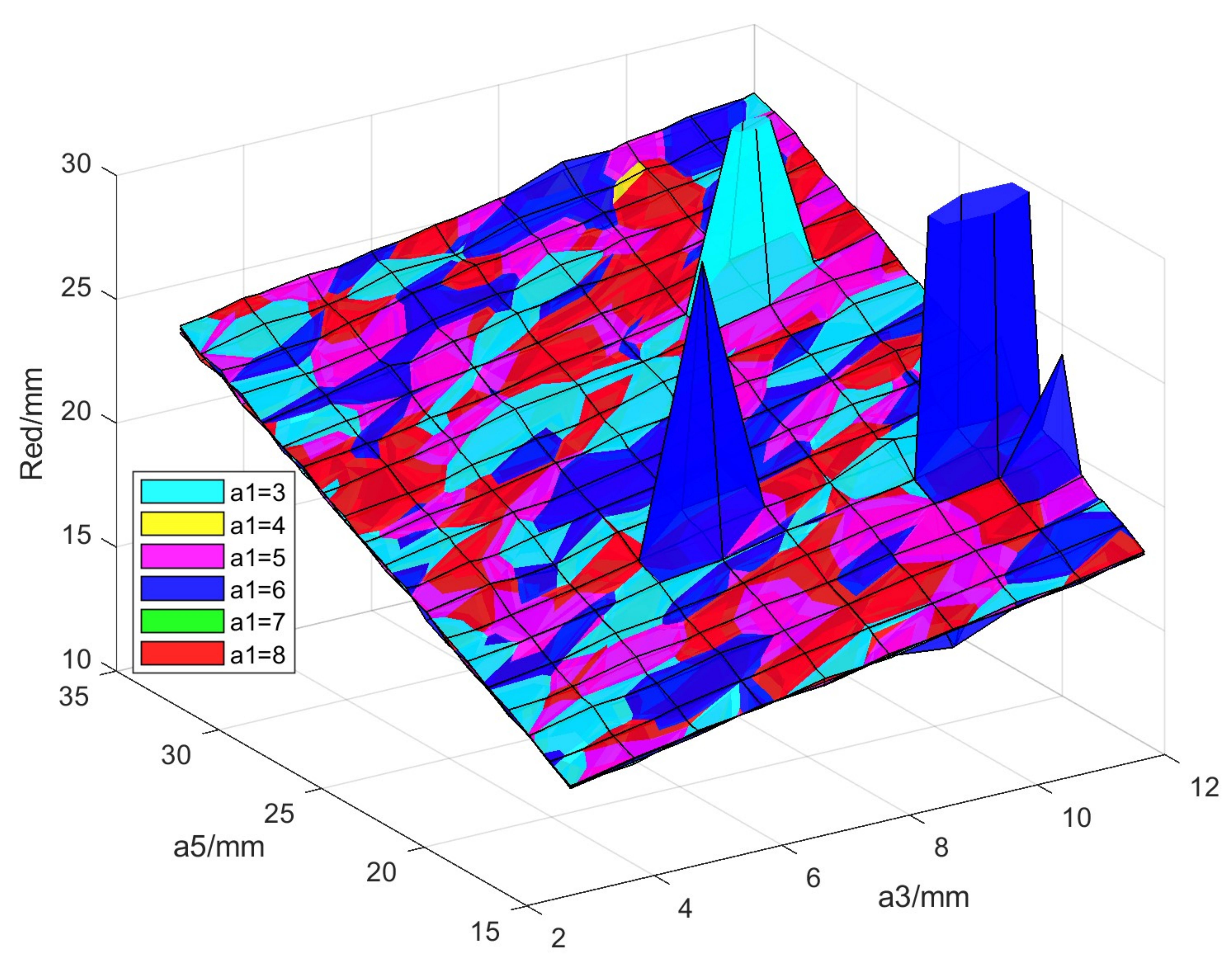}
    \vspace{-2mm}
    \caption{Stacked results of the full exploration of feasible link parameters.}
    \label{fig.OptWSRes}
    \vspace{-4mm}
\end{figure}

On the other hand, we want the total size of the miniaturized instrument to be as small as possible, or not necessarily the smallest but smaller is more applicable in confined surgical in-sites. Thus, without the simplification as stated in the preliminary solution above, we can further solve the problem step by step following the algorithm shown in Fig.\ref{fig.Optworkflow}. With much more amount of computing, the final results of volume and the equivalent radius $R_{ed}$ are also obtained. With all the exploration given different $a_1$, the stacked results of $R_{ed}$ are shown in Fig.\ref{fig.OptWSRes}.   

According to the exploration results of all link parameter combinations within the limits, we can find a series of combinations are feasible. We can choose to ensure a satisfactory dexterous workspace with equivalent redius $R_{ed} \geq 17mm$ but with a small total size considering the fabricability and functionality. Checking the values of the link parameters, one satisfactory set can be: $a_1=5mm,\ a_3=8mm$, and $a_5=18mm$. Its equivalent radius of the resulted dexterous workspace is $R_{ed} = 17.1mm$, and the volume of dexterous and reachable workspace are about $2.09\times{10}^{-5}\ m^3$ and $2.32\times{10}^{-5}\ m^3$  respectively.

\section{Discussion and Conclusion}

\subsection{Discussion}
In this work, the least-squares $n$th order polynomial fitting method is used for the workspace estimation. But for the slices near to the edges of the workspace, the boundary determination is greatly affected by more outlier points, like the 1st slice.
With the same $\delta_x$ and $S_m$ for all the slices, $2S_m$ sampling points have much higher density on those smaller planes near the edges. Moreover, the order $s$ of the polynomial fitting for each slice in a chosen workspace is kept the same. As a result, the accuracy of the boundary curve fitting is greatly influenced by the over-fitting problems in this case. Non-uniformly distributed partition method with adaptive $\delta_x$ or adaptive order $s$ for each slice can be further developed to improve the results.

The kinematic parameter optimization method presented in this paper is mainly applied in designing a miniature surgical instrument for miniaturized access to clinical interventions, but actually it also provides a general and direct way that can be used for robotic optimizing design with size and workspace restrictions in different applications.

\subsection{Conclusion}

This paper proposed a kinematic parameter optimization method based on dexterous workspace determination to achieve the optimized design a miniaturized surgical instrument with large enough dexterous workspace for surgical interventions.

The advanced Monte Carlo method based on Beta distribution is used to generate more uniformly distributed point clouds of the robot workspace, the relative manipulability with a given threshold is used to determine the representation of dexterous workspace, and the numerical technique of least-squares $n$th order polynomial fitting for the partitioned slice of point clouds are developed to achieve the determination and estimation of the workspace boundary and volume. To solve the optimization problem, simplification with given restrictions and full exploration approach 
are implemented as the analytical objective function with respect to the kinematic parameters is hard to address. Eventually, the optimized results of the kinematic parameters are obtained following the proposed exploration algorithm. Preliminary results from simplified computation verified the feasibility of the parametric combination optimization. To achieve smaller size and keep enough dexterous workspace for targeted object intervention, improved results are further addressed with full exploration method.

Future work will aim to improve the optimization design of the link parameters with larger database and develop more efficient algorithms with less computing time. The fitting method used to determine the boundary curves for each slice can be further updated to be less sensitive to outliers and make the polynomial orders for each set of the boundary points more adaptive to ensure higher accuracy.

\addtolength{\textheight}{-12cm}   

\balance

\bibliographystyle{IEEEtran}
\bibliography{IEEEabrv, SurgInstOpt}

\end{document}